\documentclass[letterpaper]{article} 
\usepackage[]{aaai23}  
\usepackage{times}  
\usepackage{helvet}  
\usepackage{courier}  
\usepackage[hyphens]{url}  
\usepackage{graphicx} 
\usepackage{natbib}  
\usepackage{caption} 
\usepackage{algorithm}
\usepackage{algorithmic}
\newcommand{\smallsim}{\smallsym{\mathrel}{\sim}}

\makeatletter
\newcommand{\smallsym}[2]{#1{\mathpalette\make@small@sym{#2}}}
\newcommand{\make@small@sym}[2]{%
  \vcenter{\hbox{$\m@th\downgrade@style#1#2$}}%
}
\newcommand{\downgrade@style}[1]{%
  \ifx#1\displaystyle\scriptstyle\else
    \ifx#1\textstyle\scriptstyle\else
      \scriptscriptstyle
  \fi\fi
}
\makeatother
\usepackage{multirow}

\newcommand{\blockcomment}[1]{}
\usepackage{amsfonts}       
\usepackage{newfloat}
\usepackage{listings}
\DeclareCaptionStyle{ruled}{labelfont=normalfont,labelsep=colon,strut=off} 
\floatstyle{ruled}
\newfloat{listing}{tb}{lst}{}
\floatname{listing}{Listing}
\usepackage{subcaption}

\title{Correlation Loss: Enforcing Correlation between Classification and Localization}
\author{
    Fehmi Kahraman\equalcontrib$^{,1}$, Kemal Oksuz\equalcontrib$^{,1}$, Sinan Kalkan$^\dagger$$^{,1,2}$, Emre Akbas\thanks{Equal contribution for senior authorship.}$^{,1,2}$
}
\affiliations{
  $^1$Dept. of Computer Engineering. $^2$METU Center for Robotics and Artificial Intelligence (ROMER) \\ Middle East Technical University (METU),
  Ankara, Turkey \\
    \{fehmi.kahraman\_01, kemal.oksuz, skalkan, eakbas\}@metu.edu.tr
}

\begin{document}

\maketitle

\begin{abstract}
Object detectors are conventionally trained by a weighted sum of classification and localization losses. Recent studies (e.g., predicting IoU with an auxiliary head, Generalized Focal Loss, Rank \& Sort Loss) have shown that forcing these two loss terms to interact with each other in non-conventional ways creates a useful inductive bias and improves performance. Inspired by these works, we focus on the correlation between classification and localization and make two main contributions: (i) We provide an analysis about the effects of correlation between classification and localization tasks in object detectors.  We identify why correlation affects the performance of various NMS-based and NMS-free detectors, and we devise measures to evaluate the effect of correlation and use them to analyze common detectors. (ii) Motivated by our observations, e.g., that NMS-free detectors can also benefit from correlation, we propose Correlation Loss, a novel plug-in loss function that improves the performance of various object detectors by directly optimizing correlation coefficients: E.g., Correlation Loss on Sparse R-CNN, an NMS-free method, yields $1.6$ AP gain on COCO and $1.8$ AP gain on Cityscapes dataset. Our best model on Sparse R-CNN reaches $51.0$ AP without test-time augmentation on COCO test-dev, reaching state-of-the-art. Code is available at: \url{https://github.com/fehmikahraman/CorrLoss}.
\end{abstract}

\section{Introduction}
\label{sec:Intro}

\begin{figure*}
    \centerline{
        \includegraphics[width=0.98\textwidth]{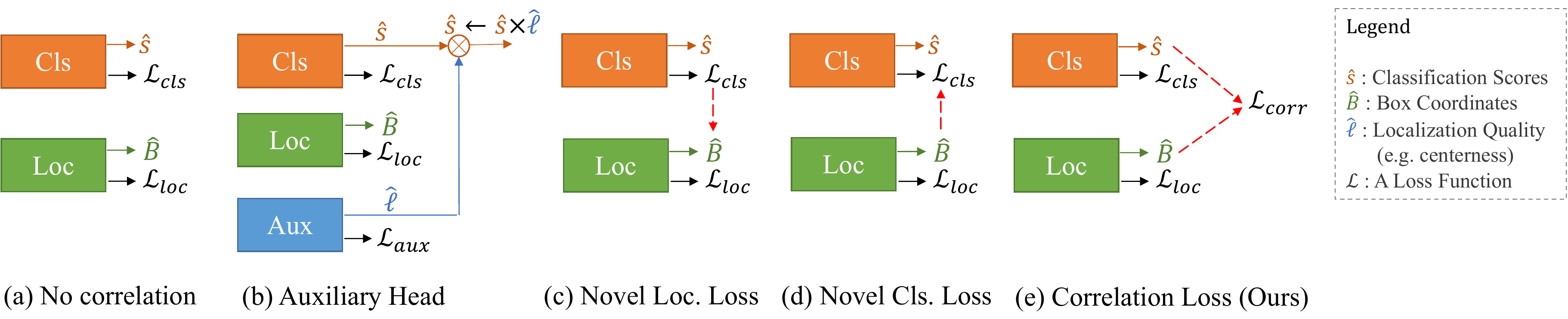}
    }
    \vspace{-1ex}
    \caption{
    Different ways of handling the classification and localization tasks from the perspective of correlation. \textbf{(a)} Conventional case of optimizing the two tasks independently (e.g., \citealt{APLoss,sparsercnn}). \textbf{(b)} An additional auxiliary head predicts centerness \cite{ATSS} or IoU  \cite{IoUNet,paa}, which introduces additional learnable parameters. \textbf{(c)} Novel loss functions replace the standard localization loss \cite{aLRPLoss} or \textbf{(d)} novel classification loss \cite{GFL, RSLoss} by more complicated ones to leverage correlation. \textbf{(e)} Our Correlation Loss explicitly optimizes a correlation coefficient. It is a simple, plug-in loss function which does not introduce additional parameters and has the flexibility to supervise classification or localisation head as well as both. Black and colored arrows respectively denote the loss functions (i.e., during training) \& the network outputs (i.e., during inference).
    \label{fig:Teaser}
} 
    \vspace{-1ex}
\end{figure*}

Most object detectors optimize a weighted sum of classification and localization losses during training. Results from recent work suggest that  performance improves when these two loss functions are forced to interact with each other in non-conventional ways as illustrated in Fig. \ref{fig:Teaser}.  
For example, training an auxiliary (aux.) head to regress the localization qualities of the positive examples, e.g. centerness, IoU or mask-IoU, has proven useful \cite{IoUNet,paa,FCOS,ATSS} (Fig. \ref{fig:Teaser}(b)). Other methods remove such auxiliary heads and aim directly to enforce correlation\footnote{In the rest of the paper, ``correlation'' will refer to the correlation between classification scores and IoUs.} in the classification or localization task during training; e.g.,  Average LRP Loss \cite{aLRPLoss} weighs the examples in the localization task by ranking them with respect to (wrt.) their classification scores (Fig. \ref{fig:Teaser}(c)). Using localization quality as an additional supervision signal for classification  has been more commonly adopted (Fig. \ref{fig:Teaser}(d)) \cite{GFL,RankDetNet,RSLoss,varifocalnet} in two main ways: (i) Score-based approaches aim to regress the localization qualities \cite{GFLv2,GFL,varifocalnet} in the classification score, and (ii) ranking-based approaches enforce the classifier to rank the confidence scores wrt. the localization qualities \cite{RankDetNet,RSLoss}.

\begin{figure}[ht]
        \captionsetup[subfigure]{}
        \centering
        \begin{subfigure}[b]{0.23\textwidth}
        \includegraphics[width=\textwidth]{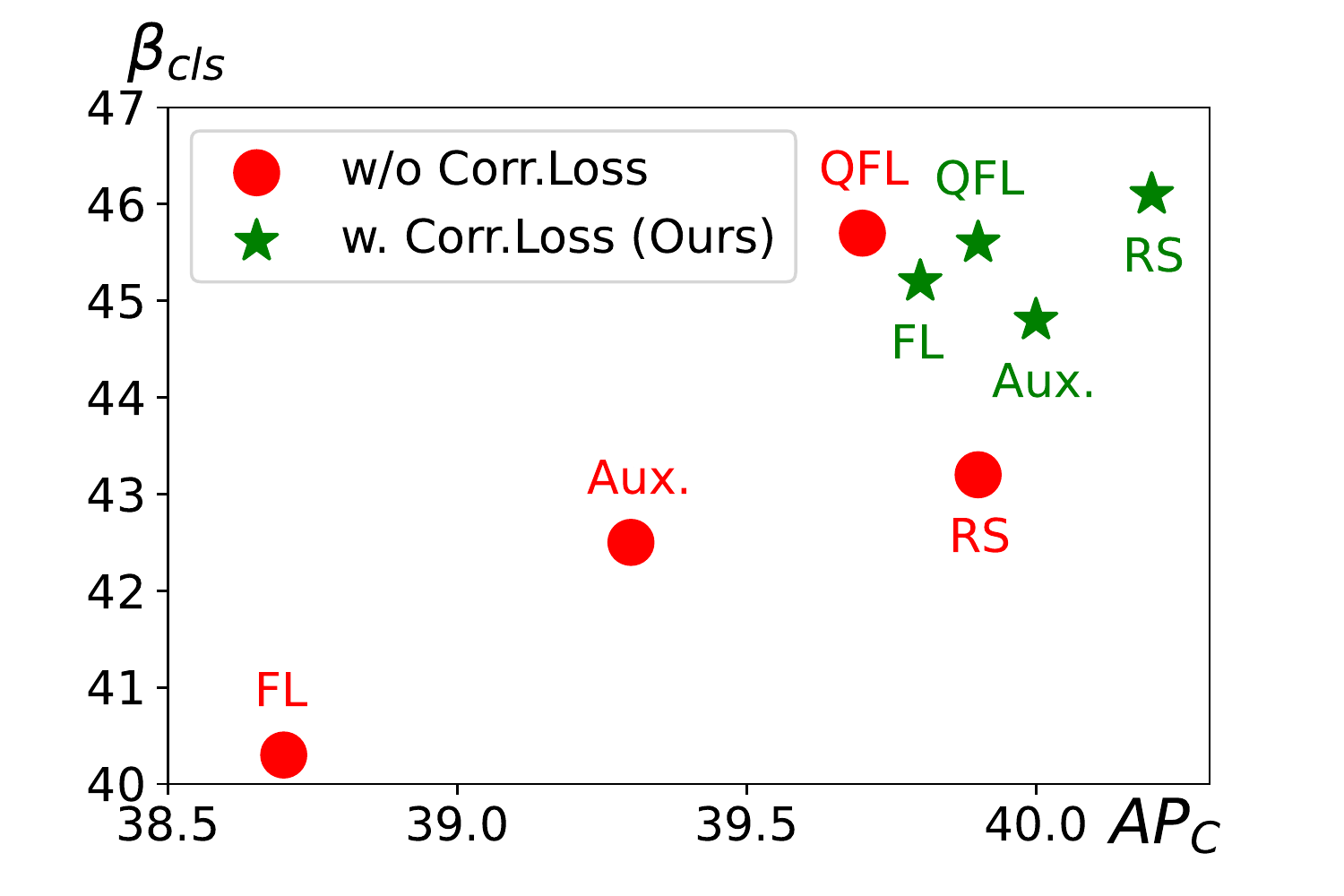}
        \caption{Detection vs. Correlation}
        \end{subfigure}
        \begin{subfigure}[b]{0.23\textwidth}
        \includegraphics[width=\textwidth]{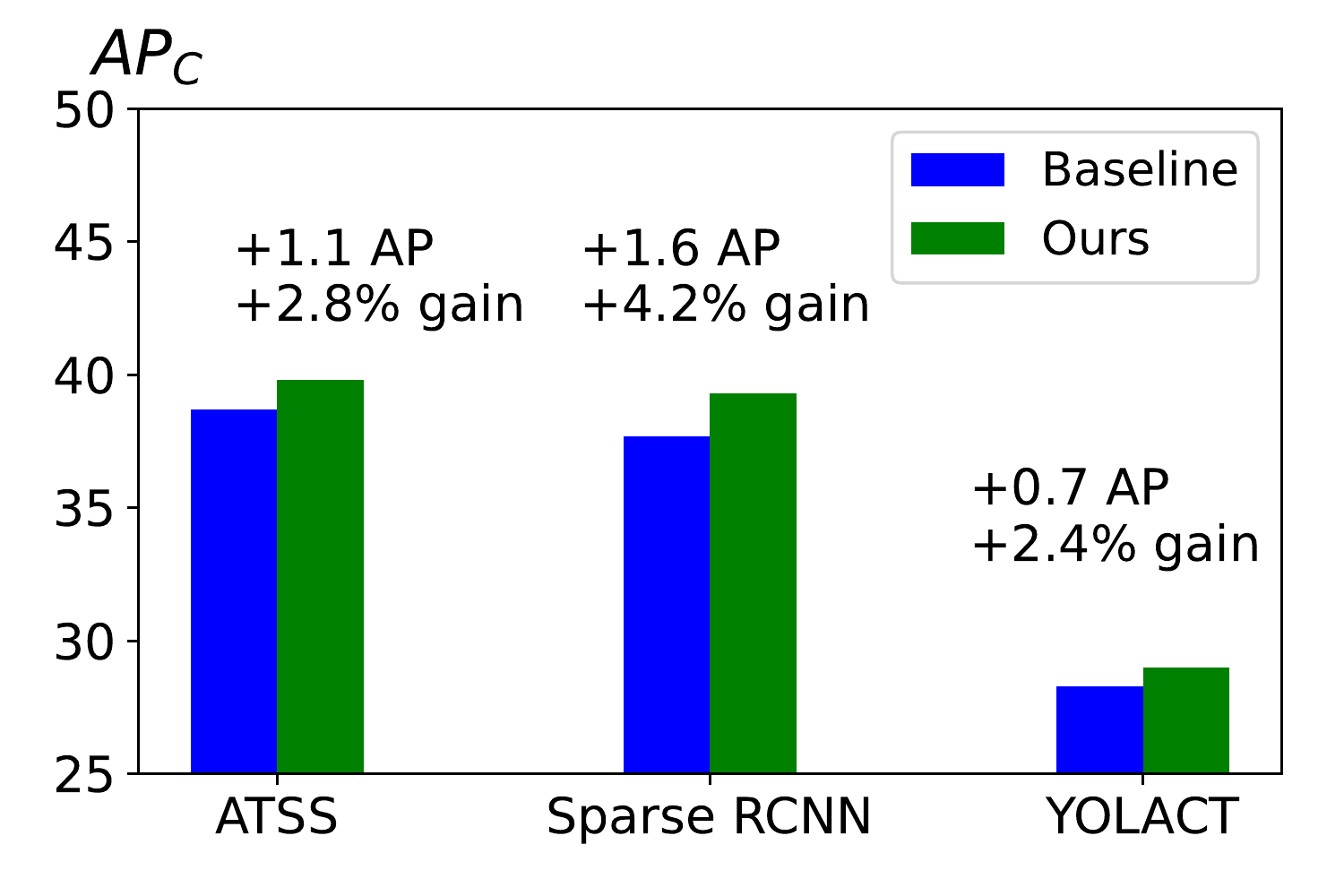}
        \caption{Effect of our Corr. Loss}
        \end{subfigure}
                \vspace{-1ex}
        \caption{\textbf{(a)} Detection performance, measured by COCO-style AP ($\mathrm{AP_C}$) vs. correlation quality, measured by class-level correlation ($\beta_{cls}$ - see Section \ref{subsec:measure} for details). The methods proposed to improve the correlation between classification and localization tasks also improve $\mathrm{AP_C}$. Compare using aux. head, QFL, RS Loss with the baseline ATSS only using Focal Loss (FL -- all in red dots) to see the positive correlation between $\mathrm{AP_C}$ and $\beta_{cls}$. Our Correlation Loss as a plug-in loss function explicitly optimizes a correlation coefficient and improves the detection performance ($\mathrm{AP_C}$) over different settings of ATSS (i.e. using FL, aux. head, QFL, RS Loss) consistently owing to increasing $\beta_{cls}$, validating our hypothesis (compare green stars with red dots).
        \textbf{(b)} Our Correlation Loss is simple-to-use and improves various methods (i) NMS-based ATSS (w/o aux. head) by $1.1 \mathrm{AP_C}$, (ii) NMS-free Sparse R-CNN by $1.6 \mathrm{AP_C}$ and (iii) YOLACT, an instance segmentation method by $0.7 \mathrm{AP_C}$. }
                \vspace{-1ex}
        \label{fig:Analysis}
\end{figure}

Improving correlation seems to have a positive effect on performance of a variety of object detectors, as shown in Fig. \ref{fig:Analysis}. However, the effect of correlation on object detectors has not been thoroughly studied. We fill this gap in this paper and first identify that correlation affects the performance of object detectors at two levels: (i) \textit{Image-level correlation}, the correlation between the classification scores and localization qualities (i.e., IoU for the rest of the paper) of the detections in a single image before post-processing, which is important to promote NMS performance, and (ii) \textit{Class-level correlation}, the correlation over the entire dataset for each class after post-processing, which is related to the COCO-style Average Precision (AP). Moreover, we quantitatively define correlation at each level to enable analyses on how well an object detector captures correlation (e.g., $\beta_{cls}$ in Fig. \ref{fig:Analysis}(a)). Then, we provide an analysis on both levels of correlation and draw important observations using common models. Finally, to better exploit  correlation,
we introduce a more direct mechanism to enforce correlation:  \textit{Correlation Loss}, a simple plug-in and detector-independent loss term (Fig. \ref{fig:Teaser}(e)), improving performance for a wide range of object detectors including NMS-free detectors, aligning with our analysis (Fig. \ref{fig:Analysis}(b)). Similar to the novel loss functions \cite{GFL,RSLoss,varifocalnet}, our Correlation Loss boosts the performance without an auxiliary head, but different from them, it is a simple plug-in technique that can easily be incorporated into any object detector, whether NMS-based or NMS-free.

Our main contributions are: \textbf{(1)} We identify how correlation affects NMS-based and NMS-free detectors, and design quantitative measures to analyze a detector wrt. correlation. \textbf{(2)} We analyze the effects of correlation at different levels on various object detectors. \textbf{(3)} We propose Correlation Loss as a plug-in  loss function to optimize correlation explicitly. Thanks to its simplicity, our loss function can be easily incorporated into a diverse set of object detectors and improves the performance of e.g., Sparse R-CNN up to $1.6$ AP and $2.0 \mathrm{AP_{75}}$, suggesting, for the first time, that NMS-free detectors can also benefit from correlation. Our best model yields $51.0$ AP, reaching state-of-the art.

\section{Background and Related Work}
\label{sec:RelatedWork}
\begin{figure*}
    \centerline{
        \includegraphics[width=0.98\textwidth]{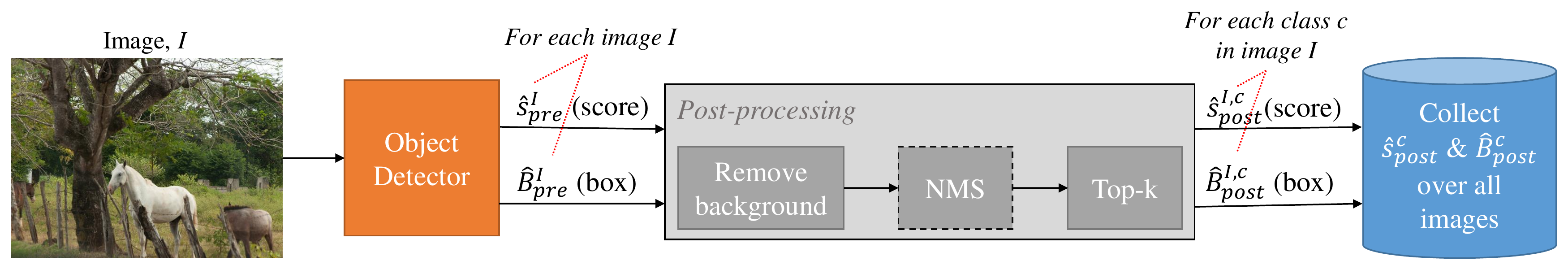}
    }
        \vspace{-1ex}
    \caption{Object detection pipeline and notation. Given an input image, $I$, NMS-based detectors yield raw detections before post-processing, each of which has a predicted bounding box (BB) and an array of confidence scores over GT classes. We denote the confidence scores and the predicted BBs pertaining to the \textit{positive detections}, i.e., the detections matching with GT objects during training, by $\hat{s}^I_{pre}$ and $\hat{B}^I_{pre}$, respectively. To obtain final detections, raw detections are post-processed in three steps: (i) Detections with low confidence scores, i.e., background, are removed, (ii) duplicates are eliminated by NMS, and (iii) top-k scoring detections are kept. As for these final detections, we denote the confidence scores and BBs of \textit{true positive detections} for class $c$ in a single image $I$ by $\hat{s}^{I,c}_{post}$ and $\hat{B}^{I,c}_{post}$ respectively, and over the entire dataset by $\hat{s}^c_{post}$ and $\hat{B}^c_{post}$. As for NMS-free detectors; NMS, dashed gray box in post-processing, is excluded, hence post-processing is lighter.
    \label{fig:Pipeline}
        \vspace{-1ex}
} 
\end{figure*}
\textbf{Object Detection Pipeline.} We group object detectors wrt. their usage of NMS (Fig. \ref{fig:Pipeline} presents overview \& notation): 

\textit{1. NMS-based Detectors.} To  detect all objects with different scales, locations and aspect ratios; most methods \cite{MaskRCNN,FoveaBox,CornerNet,FocalLoss,FasterRCNN,FCOS,ATSS} employ a large number of object hypotheses (e.g., anchors, points), which are labeled as positive (a.k.a. foreground) or negative (a.k.a. background) during training, based on whether/how they match GT boxes \cite{ATSS,FreeAnchor}. In this setting, there is no restriction for an object to be predicted by multiple object hypotheses, causing duplicates. Accordingly, during inference, NMS picks the detection with the largest confidence score among the detections that overlap more than a predetermined IoU threshold  to avoid duplicate detections.

\textit{2. NMS-free Detectors.} An emerging research direction is to remove the need for doing NMS, simplifying the detection pipeline \cite{DETR,DynamicDETR,sparsedetr,sparsercnn,OneNet,DDETR}. This is achieved by ensuring a one-to-one matching between the GTs and detections, which  supervises the detector to avoid duplicates in the first place.
    
\textbf{Methods Enforcing Correlation.} One common way to ensure correlation is to use an additional auxiliary head, supervised by the localization quality of a detection such as centerness \cite{FCOS,ATSS}, IoU \cite{IoUNet}, mask IoU \cite{maskscoring} or uncertainty \cite{KLLoss}, during training. During inference, the predictions of the auxiliary head are then combined with those of the classifier to improve detection performance. Recent methods show that the auxiliary head can be removed, and either (i) the regressor can prioritize the positive examples \cite{aLRPLoss} or (ii) the classifier can be supervised to prioritize detections with confidence scores. The latter is ensured either by regressing the IoUs by the classifier \cite{GFL,varifocalnet} or by training the classifier to rank confidence scores \cite{RankDetNet,RSLoss} wrt. IoUs. Unlike these methods, TOOD \cite{tood} takes correlation into account mainly while designing the model, particularly the detection head, i.e., not the loss function.

\textbf{Correlation Coefficients.} Correlation coefficients measure the strength and direction of the ``relation'' between two sets, $X=\{x_1, ..., x_N \}$ and $Y=\{y_1, ..., y_N \}$. Different relations are evaluated by different correlation coefficients: (i) \textit{Pearson correlation coefficient}, denoted by $\alpha(\cdot,\cdot)$, measures the linear relationship between the sets, (ii) \textit{Spearman correlation coefficient}, $\beta(\cdot,\cdot)$, corresponds to the ranking relationship and (iii) \textit{Concordance correlation coefficient}, $\gamma(\cdot,\cdot)$, is more strict, measuring the similarity of the values and maximized when $x_i = y_i$ for all $i \in {1, ..., N}$. All correlation coefficients have a range of $[-1, +1]$ where positive/negative correlation corresponds to increasing/decreasing relation, while $0$ implies no correlation between $X$ and $Y$.

\textbf{Comparative Summary.} In this paper, we comprehensively identify and analyze the effect of explicitly correlating classification and localization in object detectors. Unlike other methods that also enforce correlation, some of which are tested only on a single architecture \cite{maskscoring,IoUNet,FCOS}, we propose a simple solution by directly optimizing the correlation coefficient, which is  auxiliary-head free and easily  applicable to \textit{all} object detectors, whether NMS-based or NMS-free. Also, ours is the first to work on NMS-free detectors in this context.
\section{Effects of Correlation on Object Detectors}

This section presents why maximizing correlation is important for object detectors, introduces  measures to evaluate object detectors wrt. correlation and  provides an analysis on methods designed for improving correlation.

\begin{figure*}
    \centerline{
        \includegraphics[width=0.95\textwidth]{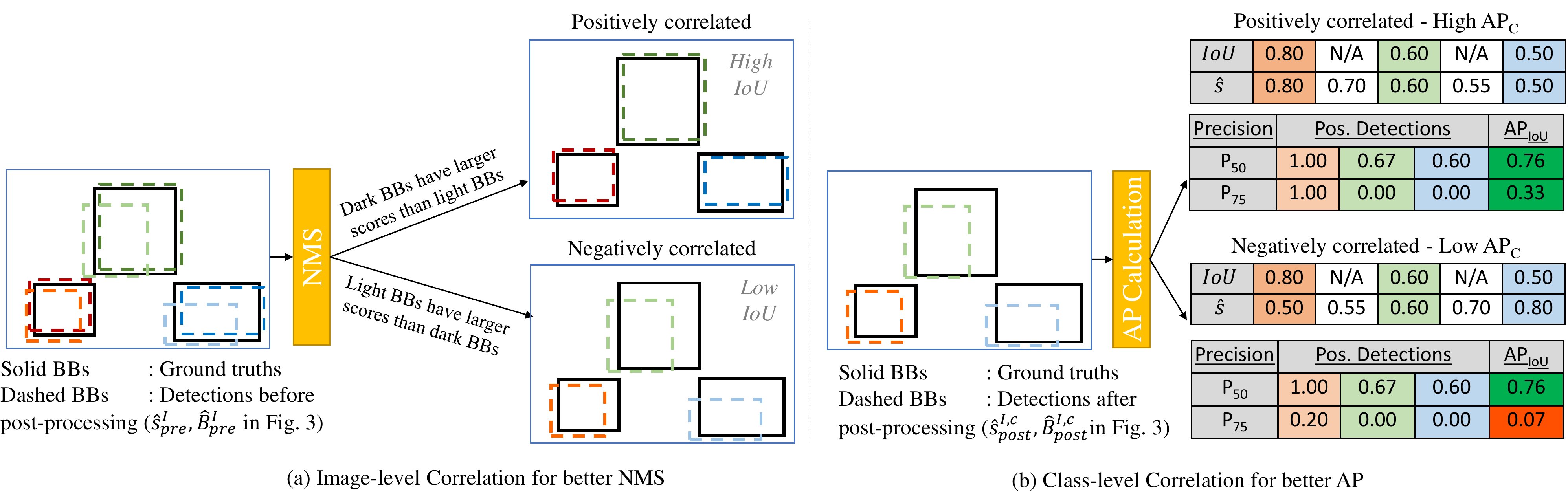}
    }
        \vspace{-1ex}
    \caption{How correlation affects detection performance. \textbf{(a)} Image-level correlation: Given detections before post-processing, NMS benefits from image-level correlation, thereby yielding detections with better IoU. Compare IoUs of detections in ``positively correlated'' (i.e., when the dark-colored ones have larger score) and ``negatively correlated'' (i.e., when the light-colored ones have larger score) outputs after NMS. \textbf{(b)} Class-level correlation: Given detections after post-processing, APs with larger IoUs and COCO-style AP benefit from positive class-level correlation
    (compare $\mathrm{AP_{IoU}}$ columns in ``positively correlated'' and ``negatively correlated'' outputs after AP Calculation to see lower $\mathrm{AP_{75}}$ for the ``negatively correlated'' output in the red cell). $\mathrm{P_{IoU}}$: Precision computed on a detection using the threshold $\mathrm{{IoU}}$, True positives are color-coded in tables and input, white cells: false positives, and hence their IoU is not available, N/A. 
    \label{fig:Types}
        \vspace{-1ex}
} 
\end{figure*}

\subsection{How Correlation Affects Object Detectors
}
\label{subsec:revisit}
\label{subsec:measure}

Detectors are affected by correlation at two levels (Fig. \ref{fig:Types}):

\textbf{Image-level Correlation.} This level of correlation corresponds to the correlation between the classification scores and IoUs of the detections in a single image before post-processing, and accordingly, we measure it with the Spearman correlation coefficient\footnote{While analyzing object detectors in terms of correlation, we employ Spearman correlation coefficient, $\beta(\cdot, \cdot)$, to measure the relation between the ranks of the values (i.e., scores and IoUs) instead of the values themselves, and aim to isolate the correlation quality from the localization and classification performances.}, $\beta(\cdot, \cdot)$, averaged over images. Denoting the set of images to be evaluated by $\mathcal{I}$ and IoUs between the BBs of the positive detections ($\hat{B}^I_{pre}$, Fig.  \ref{fig:Pipeline}) and their associated GTs by $\mathrm{IoU}^I_{pre}$, image-level correlation is measured as follows:

\begin{equation}
    \label{eq:pergt}
    \beta_{img} = \frac{1}{|\mathcal{I}|} \sum_{I \in \mathcal{I}} \beta(\mathrm{IoU}^I_{pre}, \hat{s}^I_{pre}).
\end{equation}

%
Maximizing image-level correlation is important for NMS-based detectors since NMS aims to suppress duplicates, i.e., to keep only a single detection for each GT when there is more than one. More particularly among overlapping detections (e.g., dark and light green detections in the detector output image in Fig. \ref{fig:Types}(a)), NMS picks the one with the larger score, and hence, if there is positive correlation between the confidence scores and IoUs of those overlapping detections, then the one with the best IoU (e.g., dark green detection in Fig. \ref{fig:Types}(a)) will survive and detection performance will increase.

\textbf{Class-level Correlation.} This level of correlation indicates the correlation between the classification scores and IoUs of the detections obtained after post-processing for each class. Since  class-level correlation is related to COCO-style AP, $\mathrm{AP_C}$, we average $\beta(\cdot, \cdot)$ over classes to be consistent with the computation of $\mathrm{AP_C}$:
\begin{equation}
\label{eq:acrgt}
    \beta_{cls} = \frac{1}{|\mathcal{C}|} \sum_{c \in \mathcal{C}} \beta(\mathrm{IoU}^c_{post}, \hat{s}^c_{post}),
\end{equation}
where $\mathcal{C}$ is the set of classes in the dataset and $\mathrm{IoU}^c_{post}$ is the set IoUs of BBs of true positives for class $c$ ($\hat{B}^c_{post}$, Fig. \ref{fig:Pipeline}).

Class-level correlation affects the performance of all detectors since it is directly related to $\mathrm{AP_C}$, the performance measure itself. To be more specific, $\mathrm{AP_C}$ for a single class is defined as the average of APs computed over 10 different IoU thresholds, $\mathrm{IoU} \in \{0.50, 0.55, ..., 0.95\}$, validating the true positives. For a specific  threshold $\mathrm{IoU}$, the detections are first sorted with respect to the classification scores, and then precision and recall pairs are calculated on each detection. Using these pairs, a precision-recall (PR) curve is obtained, and the area under the PR curve corresponds to the single AP value, $\mathrm{AP_{IoU}}$. When the correlation between classification and localization is maximized among true positives, larger precision values are obtained on the same detections in larger $\mathrm{IoU}$ values (e.g. $P_{75}$ of orange detection is $1.00$ and $0.20$ with positive and negative correlation respectively in Fig. \ref{fig:Types}(b)). 

\begin{table*}[t]
    \centering
    \setlength{\tabcolsep}{0.3em}
    \small

    \begin{tabular}{|l|c|c|c|c||c|c|c|c|c|c|} \hline
        &\multicolumn{4}{|c|}{Performance}& \multicolumn{6}{|c|}{Modify ranking of scores} \\ 
        \hline
        Method &$\mathrm{AP_C}$&$\mathrm{AP_{50}}$&$\mathrm{AP_{75}}$&$\beta_{img}$&$\mathrm{AP}^{-1}_C$&$\mathrm{AP}^{-1}_{50}$&$\mathrm{AP}^{-1}_{75}$&$\mathrm{AP}^{+1}_C$&$\mathrm{AP}^{+1}_{50}$&$\mathrm{AP}^{+1}_{75}$ \\ 
        \hline \hline
        \textbf{Not Enforcing Correlation}& & & & & & & & & &\\
        ATSS w. AP Loss \cite{APLoss}&$38.1$&$58.2$&$41.0$&$27.2$&$24.9$&$53.2$&$19.2$&$57.0$&$72.4$&$62.2$\\ 
        ATSS w. Focal Loss \cite{FocalLoss}&$38.7$&$57.6$&$41.5$&$27.3$&$25.6$&$51.8$&$21.1$&$55.8$&$70.6$&$60.5$\\ 
        \hline
        \textbf{Using Aux. Head}& & & & & & & & & &\\ 
        ATSS w. ctr. head \cite{ATSS}&$39.3$&$57.5$&$42.6$&$28.7$&$16.8$&$32.4$&$15.3$&$49.8$&$64.8$&$54.2$\\ 
        \hline
        \textbf{Using Novel Loss}& & & & & & & & & & \\
        ATSS w. aLRP Loss \cite{aLRPLoss} &$37.7$&$57.4$&$39.9$&$33.8$&$22.7$&$48.8$&$17.5$&$54.2$&$70.4$&$58.7$\\ 
        ATSS w. QFL \cite{GFL} &$39.7$&$58.1$&$42.7$&$33.2$&$25.7$&$51.1$&$21.9$&$55.8$&$70.9$&$60.6$\\ 
        ATSS w. RS Loss \cite{RSLoss} &$39.9$&$58.9$&$42.6$&$30.9$&$26.2$&$53.9$&$21.3$&$57.1$&$71.8$&$62.1$\\ 
        \hline
    \end{tabular}
        \vspace{-1ex}
    \caption{Evaluation of NMS-based detectors in terms of image-level correlation. See Eq. \ref{eq:pergt} for $\beta_{img}$. $\mathrm{AP}^{+1}_{\mathrm{IoU}}$ and $\mathrm{AP}^{-1}_{\mathrm{IoU}}$ refer to the upper \& lower bound APs (see analysis setup for details). The values are in \%. Our $\beta_{img}$ captures  correlation consistently, e.g. that (i) Focal Loss is improved by ctr. head and QFL and (ii) AP Loss is improved by aLRP Loss and RS Loss wrt. $\beta_{img}$. Also, there is still room for improvement for object detectors wrt. $\beta_{img}$ with a range between $27.2 \%$ and $33.8 \%$.}
        \vspace{-1ex}
    \label{tab:pergt}
\end{table*}

\begin{table*}[t]
    \centering
    \setlength{\tabcolsep}{0.3em}
    \small
    
    \begin{tabular}{|l|c|c|c|c|c|c|c|c|c|c|} \hline
        &\multicolumn{4}{|c|}{Performance}& \multicolumn{6}{|c|}{Modify ranking of scores} \\ 
        \hline
        Method &$\mathrm{AP_C}$&$\mathrm{AP_{50}}$&$\mathrm{AP_{75}}$&$\beta_{cls}$&$\mathrm{AP}^{-1}_C$&$\mathrm{AP}^{-1}_{50}$&$\mathrm{AP}^{-1}_{75}$&$\mathrm{AP}^{+1}_C$&$\mathrm{AP}^{+1}_{50}$&$\mathrm{AP}^{+1}_{75}$\\ 
        \hline\hline
         \textbf{Not Enforcing Correlation}& & & & & & & & & &\\
         \textit{- NMS-free Detectors}& &&  & & & & & & & \\
        \ \ Sparse R-CNN \cite{sparsercnn} &$37.7$ &$55.8$&$40.5$&$37.5$ &$30.1$ &$55.8$ & $28.9$ &$48.6$ &$55.8$ & $52.7$ \\ 
        \ \ DETR \cite{DETR}& $40.1$ & $60.6$&$42.0$&$47.0$ & $32.9$ &$60.6$ &$30.6$ & $51.9$ & $60.6$ &$55.8$ \\ 
        \textit{- NMS-based Detectors}& &&  & & & & & & & \\
        \ \ ATSS w. AP Loss \cite{APLoss}&$38.1$&$58.2$&$41.0$&$39.4$ &$30.0$ &$58.2$ &$26.6$ &$48.5$ &$58.2$ &$54.0$ \\ 
        \ \ ATSS w. Focal Loss \cite{FocalLoss}&$38.7$&$57.6$&$41.5$&$40.3$ &$30.2$ &$57.6$ &$27.3$ &$48.7$ &$57.6$ &$53.6$\\ 
        \hline
        \textbf{Using Aux. Head}& & & & & & & & & & \\ 
        ATSS w. ctr. head \cite{ATSS} &$39.3$&$57.4$&$42.5$&$42.5$ &$30.2$ &$57.4$ &$27.6$ &$48.7$ &$57.4$ &$53.5$\\ 
        \hline
        \textbf{Using Novel Loss}& & & & & & & & & &\\
        ATSS w. aLRP Loss \cite{aLRPLoss} &$37.7$&$57.4$&$39.9$&$42.0$ &$29.1$ &$57.4$ &$25.0$ &$47.8$ &$57.4$ &$52.7$ \\ 
        ATSS w. QFL \cite{GFL}&$39.7$&$58.1$&$42.7$&$45.7$ &$30.6$ &$58.1$ &$27.7$ &$49.1$ &$58.1$ &$53.9$ \\ 
        ATSS w. RS Loss \cite{RSLoss}&$39.9$&$58.9$&$42.6$&$43.2$ &$31.1$ &$58.9$ &$28.1$ &$49.8$ &$58.9$ & $54.8$ \\
        \hline
    \end{tabular}
        \vspace{-1ex}
\caption{Evaluation of detectors wrt. class-level correlation. See Eq. \ref{eq:acrgt} for $\beta_{cls}$. $\mathrm{AP}^{+1}_{\mathrm{IoU}}$ \& $\mathrm{AP}^{-1}_{\mathrm{IoU}}$ denote upper \& lower bound APs (analysis setup for details). Values are in \%. NMS-free detectors can also benefit from class-level correlation (compare $\mathrm{AP}^{+1}_{\mathrm{C}}$ with $\mathrm{AP}_{\mathrm{C}}$ for Sparse R-CNN), and as in $\beta_{img}$ (c.f. Table \ref{tab:pergt} and its caption), $\beta_{cls}$ measures the correlation consistently. $\mathrm{AP}^{+1}_{50} = \mathrm{AP}^{-1}_{50}$ = $\mathrm{AP}_{50}$ since only modifying TPs validated from IoU=0.50 does not effect $\mathrm{AP}_{50}$ (see Fig. \ref{fig:Types}(b) for an example).}   
\vspace{-1ex}
\label{tab:acrossgt}
\end{table*}

\subsection{Analyses of Object Detectors wrt. Correlation}
\label{subsec:analyse}


\textbf{Dataset and Implementation Details.} Unless otherwise specified; we (i) employ the widely-used COCO dataset \cite{COCO} by training the models on \textit{trainval35K} (115K images), testing on \textit{minival} (5k images), comparing with SOTA on \textit{test-dev} (20k images), (ii) build upon the mmdetection framework \cite{mmdetection}, (iii) rely on AP-based measures and also use Optimal LRP (oLRP) \cite{LRPPAMI}, $\beta_{img}$ (Eq. \ref{eq:pergt}) and $\beta_{cls}$ (Eq. \ref{eq:acrgt}) to provide more insights, (iv)  keep the standard configuration of the models, (v) use a ResNet-50 backbone with FPN \cite{FeaturePyramidNetwork}, (vi) train models on 4 GPUs (A100 or V100 type GPUs) with 4 images on each GPU (16 batch size). 

\textbf{Analysis Setup.} We conduct experiments to analyze the effects of the image-level ($\beta_{img}$ -- Table \ref{eq:pergt}) and class-level ($\beta_{cls}$ -- Table \ref{eq:acrgt}) correlations. For both analyses, we compare three sets of methods, all of which are incorporated into the common ATSS baseline \cite{ATSS} (see Sec. \ref{sec:RelatedWork} for a discussion of these methods): (i) AP Loss and Focal Loss as methods not enforcing correlation, (ii) using an auxiliary head to enforce correlation, and (iii)  Quality Focal Loss (QFL), aLRP Loss and Rank \& Sort Loss as recent loss functions enforcing correlation. In our class-level analysis, we also employ NMS-free methods to demonstrate the effects of correlation on that approach. 

We compare the methods based on (i) their AP-based performance, (ii) our proposed measures for correlation (Eqs. \ref{eq:pergt} and \ref{eq:acrgt}),  and finally (iii) lower/upper bounds, $\mathrm{AP}^{+1}_{\mathrm{C}}$/$\mathrm{AP}^{-1}_{\mathrm{C}}$, obtained by modifying the ranking of the confidence scores pertaining to the GT classes of the positive detections to minimize/maximize Eq. \ref{eq:pergt} in Table \ref{eq:pergt} and Eq. \ref{eq:acrgt} in Table \ref{eq:acrgt}. More particularly, in Table \ref{eq:pergt}, given $\hat{s}^I_{pre}$ and $\hat{B}^I_{pre}$ (Fig. \ref{fig:Pipeline}), we collect the GT class probabilities of positive detections and change their ranking in $\hat{s}^I_{pre}$ within an image following the ranking order of IoUs (computed using $\hat{B}^I_{pre}$), and in Table \ref{eq:acrgt}, we do the same operation class-wise for true positives given $\hat{s}^c_{post}$ and $\hat{B}^c_{post}$ (Fig. \ref{fig:Pipeline}).
To decouple other types of errors as much as possible; in Table \ref{eq:pergt}, we \textit{do not modify} the scores of the negative detections, the predicted BBs and the scores of non-GT classes of the positive detections, and in Table \ref{eq:acrgt}, we \textit{do not modify} the scores of the false positives and the predicted BBs of the true positives. Note that achieving the upper bound in (iii) for image-level correlation also corresponds to perfectly minimizing RS Loss.

\textbf{Observations.} We observe in Tables \ref{eq:pergt} and \ref{eq:acrgt} that:

\textit{(1) Our proposed measures in Eqs. \ref{eq:pergt} and \ref{eq:acrgt} can measure the improvements in correlation consistently.} In Tables \ref{tab:pergt} and \ref{tab:acrossgt}, (i) aLRP Loss and RS Loss are proposed to improve AP Loss and (ii) aux. head and QFL are proposed to improve Focal Loss. In both tables, the proposed methods are shown to improve their baselines in terms of $\beta_{img}$ and $\beta_{cls}$, suggesting that our measures can consistently evaluate image-level and class-level correlations respectively.

\textit{(2) NMS-free detectors can also potentially benefit from correlation.} All detectors, including NMS-free ones, can exploit class-level correlation (compare $\mathrm{AP_C}$ and $\mathrm{AP}^{+1}_{C}$ to see $\sim 10$ points gap in Table \ref{tab:acrossgt}). Still, existing methods do not enforce this correlation on NMS-free detectors.

\textit{(3) Existing methods enforcing correlation have still a large room for improvement.} Considering that $\beta_{img} \in [27.2 \%, 33.8 \%]$ (Table \ref{tab:pergt}) and $\beta_{cls} \in [37.5 \%, 47.0\%]$ (Table \ref{tab:acrossgt}), there is still room for improvement wrt. correlation.



\textit{(4) While significantly important, improving correlation may not always imply performance improvement.} For example, aLRP Loss in Table \ref{tab:pergt} has the largest correlation but the lowest $\mathrm{AP_C}$. Such a situation can arise, for example, when a method does not have  good localization performance. In the extreme case, assume a detector yields perfect $\beta_{img}$, image-level ranking correlation, but the IoUs of all positive examples are less than $0.50$ implying no TP at all. Hence, boosting the correlation, while simultaneously preserving a good performance in each branch, is critical. 

\section{Correlation Loss: A Novel Loss Function for Object Detection}
\label{sec:CorrelationLoss}

 Correlation (Corr.) Loss is a simple plug-in loss function to improve correlation of classification and localization tasks. Correlation Loss is unique in that it can be easily incorporated into any object detector, whether NMS-based or NMS-free (see Observation (2) - Sec. \ref{subsec:analyse}), and improves  performance without affecting the model size, inference time and with negligible effect on training time (Sec. \ref{subsec:ablation}). Furthermore, from a fundamental perspective, Corr. Loss can supervise both of the classification and localisation heads for a better correlation while existing methods generally focus on a single head such as classification (Fig. \ref{fig:Teaser}).

\textbf{Definition.} Given an object detector with loss function $\mathcal{L}_{OD}$, our Correlation Loss ($\mathcal{L}_{corr}$) is simply added using a weighting hyper-parameter $\lambda_{corr}$:  
\begin{equation}
    \label{eq:LossCorr}
    \mathcal{L}_{OD}+ \lambda_{corr} \mathcal{L}_{corr}.
\end{equation}
$\mathcal{L}_{corr}$ is the Correlation Loss defined as: 
\begin{equation}
    \label{eq:LossCorr2}
    \mathcal{L}_{corr} = 1 - \rho(\hat{\mathrm{IoU}}, \hat{\mathrm{s}}),
\end{equation}
where $\rho(\cdot, \cdot)$ is a correlation coefficient;  $\hat{\mathrm{s}}$ and $\hat{\mathrm{IoU}}$ are the confidence scores of the GT class and IoUs of the predicted BBs pertaining to the positive examples in the batch. 

\textbf{Practical Usage.} To avoid promoting  trivial cases with high correlation but low performance (Observation (4) - Sec. \ref{subsec:analyse}), similar to QFL \cite{GFL} and RS Loss \cite{RSLoss}, we only use the gradients of $\mathcal{L}_{corr}$ wrt. classification score, i.e., we backpropagate the gradients through only the classifier. We mainly adopt two different correlation coefficients for $\rho(\cdot, \cdot)$ and obtain two versions of Correlation Loss: (i) \textit{Concordance Loss}, defined as the Correlation Loss when Concordance correlation coefficient is optimized ($\rho(\cdot, \cdot)=\gamma(\cdot, \cdot)$), which aims to match the confidence scores with IoUs. (ii) \textit{Spearman Loss} as Correlation Loss when Spearman correlation coefficient is optimized ($\rho(\cdot, \cdot)=\beta(\cdot, \cdot)$), thereby enforcing the ranking of the classification scores considering IoUs. To tackle the non-differentiability of ranking operation while computing Spearman Loss, we leverage the differentiable sorting operation from Blondel et al. \cite{softsort}. When applying our Correlation Loss to NMS-free methods, which use an iterative multi-stage loss function, we incorporate $\mathcal{L}_{corr}$ to every stage. 

\section{Experimental Evaluation}
\label{sec:experiments}

We evaluate Corr. Loss on (i) the COCO dataset with five different object detectors of various types (Sparse R-CNN as NMS-free, FoveaBox as anchor-free, RetinaNet as anchor-based, ATSS and PAA using auxiliary head),  and one instance segmentation method, YOLACT; and (ii) an additional dataset (Cityscapes)  for the method with the largest gain, i.e., Sparse R-CNN.



\subsection{Comparison with Methods Not Considering Correlation}
We train these five object detectors and the instance segmentation method  (Tables \ref{tab:nmseval} and \ref{tab:instancesegmentation}) with and without our Corr. Loss (as Concordance Loss or Spearman Loss).

\begin{table}
    \centering
    \small
    \setlength{\tabcolsep}{0.05em}
    \begin{tabular}{|c|l|c|c|c||c|} \hline
     \phantom{De}   & \multicolumn{1}{|c|}{Method} & $\mathrm{AP_C} \uparrow$ & $\mathrm{AP_{50}} \uparrow$ & $\mathrm{AP_{75}} \uparrow$ & $\mathrm{oLRP} \downarrow$ \\ \hline\hline
    \parbox[t]{2mm}{\multirow{12}{*}{\rotatebox[origin=c]{90}{\textit{\scriptsize NMS-based}}}} & Retina Net {\scriptsize \cite{FocalLoss}}&$36.5$&$55.4$&$39.1$&$70.7$\\
    & \quad w. Conc.Corr (Ours) &$37.0$&$\mathbf{55.7}$&$39.7$ &$70.2$ \\ 
    & \quad w. Spear.Corr (Ours) &$\mathbf{37.5}$&$55.4$&$\mathbf{40.5}$& $\mathbf{69.7}$ \\ \cline{2-6}
    & Fovea Box {\scriptsize \cite{FoveaBox}}&$36.4$&$\mathbf{56.5}$&$38.6$&$70.2$ \\
    & \quad w. Conc.Corr (Ours) &$\mathbf{37.1}$&$56.4$&$\mathbf{39.6}$&$\mathbf{69.7}$\\ 
    & \quad w. Spear.Corr (Ours) &$37.0$&$55.6$&$39.3$& $70.0$\\ \cline{2-6}
    & ATSS {\scriptsize \cite{ATSS}} &$38.7$&$57.6$&$41.5$&$69.0$ \\
    & \quad w. Conc.Corr (Ours)&$\mathbf{39.8}$&$\mathbf{57.9}$&$\mathbf{43.2}$&$\mathbf{68.2}$ \\ 
    & \quad w. Spear.Corr (Ours) &$39.3$&$56.6$&$42.5$&$68.7$  \\ \cline{2-6}
    & PAA {\scriptsize \cite{paa}} &$39.9$&$57.3$&$43.4$ &$68.6$ \\
    & \quad w. Conc.Corr (Ours) &$\mathbf{40.7}$&$\mathbf{58.8}$&$\mathbf{44.3}$&$\mathbf{67.7}$ \\ 
    & \quad w. Spear.Corr (Ours) &$40.4$&$58.0$&$43.7$&$67.8$
\\   \hline \hline
    \multirow{3}{*}{\rotatebox[origin=c]{90}{\parbox[c]{1cm}{\centering\textit{\scriptsize NMS-free}}}}&      Sparse R-CNN {\scriptsize \cite{sparsercnn}}&$37.7$&$55.8$&$40.5$&$69.5$\\
    & \quad w. Conc.Corr (Ours) &$38.9$&$\mathbf{57.2}$&$41.8$&$\mathbf{68.1}$ \\ 
    & \quad w. Spear.Corr (Ours) &$\mathbf{39.3}$ &$56.7$ &$\mathbf{42.5}$ &$68.3$\\ \hline
    \end{tabular}
                \vspace{-1ex}
    \caption{Comparison on detectors not considering correlation. Accordingly, we remove aux. heads from ATSS \cite{ATSS} and PAA \cite{paa} for fair comparison (see Table \ref{tab:enforceCorr} for comparison with aux. heads and novel loss functions). We use ResNet-50 and train the models for 12 epochs. Simply incorporating our Corr. Loss provides (i) $\smallsim 1 \mathrm{AP_C}$ improvement for NMS-based detectors consistently and (ii) $\smallsim 1.5 \mathrm{AP_C}$ on the NMS-free Sparse R-CNN.}
                \vspace{-1ex}
    \label{tab:nmseval}
\end{table}

\textbf{NMS-based Detectors.} Table \ref{tab:nmseval} suggests $\smallsim 1.0 \mathrm{AP_C}$ gain on NMS-based detectors: (i) Spearman Loss ($\lambda_{corr}=0.1$) improves RetinaNet by $1.0 \mathrm{AP_C}$ and $\mathrm{oLRP}$, (ii) Concordance Loss ($\lambda_{corr}=0.2$) enhances anchor-free FoveaBox by 0.7$\mathrm{AP_C}$, and (iii) Concordance Loss ($\lambda_{corr}=0.3$) improves ATSS   and PAA by $\smallsim 1 \mathrm{AP_C}$ and $\smallsim 1 \mathrm{oLRP}$. 

\textbf{NMS-free Detectors.} Our results in Table \ref{tab:nmseval} suggest that Sparse R-CNN,  an NMS-free method, can also benefit from our Corr. Loss: (i) Both Concordance ($\lambda_{corr}=0.3$) and Spearman Losses ($\lambda_{corr}=0.2$) improve baseline; (ii) Spearman Loss improves $\mathrm{AP_C}$ significantly by up to $1.6$; (iii) as hypothesized, the gains are owing to APs with larger IoUs, e.g., $\mathrm{AP_{75}}$ improves by up to $2.0$, and (iv) gains persist in a stronger setting of Sparse R-CNN (Appendix).


\textbf{Cityscapes dataset.} To see the effect of Corr. Loss over different scenarios, we train Sparse R-CNN with Spearman Loss (the model that has the best gain over baseline in Table \ref{tab:nmseval}), on the Cityscapes dataset \cite{Cityscapes} ($\lambda_{corr}=0.6$), a dataset for autonomous driving object detection. Table \ref{tab:cityscapes} presents that (i) Spearman Loss also improves baseline Sparse R-CNN on Cityscapes by $1.8$ AP and (ii) our gain mainly originates from APs with larger IoUs, i.e. $\mathrm{AP_{75}}$ improves by more than $3$ points, from $37.6$ to $40.8$.

\textbf{Instance Segmentation.} We train YOLACT  \cite{yolact} as an instance segmentation method with Corr. Loss and observed $0.7$ mask AP gain using Spearman Loss ($\lambda_{corr}=0.5$ - Table \ref{tab:instancesegmentation}), implying $1.7\%$ relative gain.

\begin{table}
    \centering
    \footnotesize
    
    \begin{tabular}{|c|c|c|c|} \hline
    Method&$\mathrm{AP}$&$\mathrm{AP_{50}}$&$\mathrm{AP_{75}}$ \\ \hline
    Sparse R-CNN &$39.0$&$63.1$&$37.6$\\
     w. Spear.Corr (Ours)  & $\mathbf{40.8}$ &$\mathbf{64.4}$&$\mathbf{40.8}$ \\ \hline
    \end{tabular}
                \vspace{-1ex}
    \caption{Results on Cityscapes dataset.}
                \vspace{-1ex}
    \label{tab:cityscapes}
\end{table}

\blockcomment{
\begin{table}
    \centering
    \small
    \setlength{\tabcolsep}{0.7em}
    
    \begin{tabular}{|c|c|c|c|} \hline
    Method&$\mathrm{AP}$&$\mathrm{AP_{50}}$&$\mathrm{AP_{75}}$ \\ \hline
    Sparse R-CNN \cite{sparsercnn} &$45.0$&$64.1$&$48.9$\\
     w. Conc.Corr (Ours)&$45.5$&$\mathbf{64.4}$&$49.7$\\
     w. Spear.Corr (Ours)&$\mathbf{46.1}$ &$64.0$&$\mathbf{50.4}$\\ \hline
    \end{tabular}
    \caption{Comparison with stronger Sparse R-CNN.}
    \label{tab:strongersetting}
\end{table}
}
\begin{table}
    \centering
    \setlength{\tabcolsep}{0.1em}
    \footnotesize
    
    \begin{tabular}{|c|c|c|c|} \hline
    Method&$\mathrm{AP_C^{mask}}$&$\mathrm{AP_{50}^{mask}}$&$\mathrm{AP_{75}^{mask}}$ \\ \hline\hline
    YOLACT \cite{yolact}&$28.3$&$47.8$&$28.8$\\
     w. Conc.Corr (Ours) &$28.8$&$\mathbf{48.3}$&$29.6$ \\
     w. Spear.Corr (Ours)  & $\mathbf{29.0}$ &$48.3$&$\mathbf{30.0}$ \\ \hline
    \end{tabular}
                \vspace{-1ex}
    \caption{Comparison with YOLACT.}
    \label{tab:instancesegmentation}
                \vspace{-1ex}
\end{table}

\blockcomment{
\begin{table}
    \setlength{\tabcolsep}{0.3em}
    \footnotesize
    \caption{Effect of Corr.Loss on YOLACT.
    }
    \label{tab:instancesegmentation}
    \begin{tabular}{|c|c|c|c|} \hline
    Method&$\mathrm{AP_C^{mask}}$&$\mathrm{AP_{50}^{mask}}$&$\mathrm{AP_{75}^{mask}}$ \\ \hline
    YOLACT \cite{yolact}&$28.3$&$47.8$&$28.8$\\
     w. Conc.Corr (Ours) &$28.8$&$\mathbf{48.3}$&$29.6$ \\
     w. Spear.Corr (Ours)  & $\mathbf{29.0}$ &$48.3$&$\mathbf{30.0}$ \\ \hline
    \end{tabular}
\end{table}
}

\begin{table}
    \centering
    \setlength{\tabcolsep}{0.05em}
    \small
    
    \begin{tabular}{|c|c|c|c||c|c|c||c||c|c|} \hline
     Aux.&QFL& RS Loss &Ours&$\mathrm{AP_{C}}$&$\mathrm{AP_{50}} $&$\mathrm{AP_{75}} $&oLRP $\downarrow$&$\beta_{img} \uparrow$&$\beta_{cls} \uparrow$\\ \hline\hline
     & & & & $38.7$&$57.6$&$41.5$&$68.9$&$27.3$&$40.3$\\
     \checkmark& & & &$39.3$&$57.5$&$42.6$&$68.6$&$28.7$&$42.5$\\ 
    &\checkmark& & &$39.7$&$58.1$&$42.7$&$68.0$&$33.2$&$45.7$\\
    & & \checkmark & &$39.9$&$\mathbf{58.9}$&$42.6$&$67.9$&$30.9$&$43.2$\\\hline
    & &  &\checkmark&$39.8$&$57.6$&$43.1$&$68.2$&$31.6$&$45.2$ \\ 
    \checkmark& &  &\checkmark&$40.0$&$58.0$&$43.3$&$68.0$&$31.1$&$44.8$\\
    & \checkmark&  &\checkmark&$39.9$&$58.2$&$43.2$&$\mathbf{67.7}$&$\mathbf{34.6}$&$45.6$\\
    & &\checkmark &\checkmark&$\mathbf{40.2}$&$58.6$&$\mathbf{43.5}$&$67.9$&$33.6$&$\mathbf{46.1}$\\\hline 
    \end{tabular}
                \vspace{-1ex}
    \caption{Comparison with methods enforcing correlation. Corr. Loss (i) reaches similar results with existing methods on ATSS, (ii) is complementary to those methods thanks to its simple design and (iii) once combined with RS Loss, outperforms compared methods.}
                \vspace{-1ex}
    \label{tab:enforceCorr}
\end{table}

\begin{table*}[ht]
    \centering
    \setlength{\tabcolsep}{0.15em}
    \small
    
    \begin{tabular}{|c|c|c|c||c|c|c|c|c|c|c|} \hline
    &Method&Backbone&Epochs&$\mathrm{AP_{C}}$&$\mathrm{AP_{50}}$&$\mathrm{AP_{75}}$&$\mathrm{AP_{S}}$&$\mathrm{AP_{M}}$&$\mathrm{AP_{L}}$&Venue\\ \hline\hline  
    \multirow{6}{*}{\rotatebox{90}{\scriptsize{\textit{NMS-based}}}}&ATSS \cite{ATSS}&ResNet-101-DCN&24&$46.3$&$64.7$&$50.4$&$27.7$&$49.8$&$58.4$&CVPR 2020\\
    &GFLv2 \cite{GFLv2}&ResNet-101-DCN&24&$48.3$&$66.5$&$52.8$&$28.8$&$51.9$&$60.7$&CVPR 2021\\
    &aLRP Loss \cite{aLRPLoss}&ResNeXt-101-DCN &100&$48.9$&$69.3$&$52.5$&$30.8$&$51.5$&$62.1$&NeurIPS 2020\\
    &VFNet \cite{varifocalnet}&ResNet-101-DCN &24&$49.2$&$67.5$&$53.7$&$29.7$&$52.6$&$62.4$&CVPR 2021\\
    &DW \cite{DW}&ResNet-101-DCN&24&$49.3$&$67.6$&$53.3$&$29.2$&$52.2$&$63.5$&CVPR 2022\\
    &TOOD \cite{tood}&ResNet-101-DCN &24&$49.6$&$67.4$&$54.1$&$30.5$&$52.7$&$62.4$&ICCV 2021\\
    &RS-Mask R-CNN+ \cite{RSLoss}&ResNeXt-101-DCN &36&$50.2$&$\mathbf{70.3}$&$54.8$&$\mathbf{31.5}$&$53.5$&$63.9$&ICCV 2021\\    
    \hline
    \multirow{4}{*}{\rotatebox{90}{\scriptsize{\textit{NMS-free}}}}&TSP R-CNN \cite{TSPRCNN}&ResNet-101-DCN&96&$47.4$&$66.7$&$51.9$&$29.0$&$49.7$&$59.1$&ICCV 2021\\
    &Sparse R-CNN \cite{sparsercnn}&ResNeXt-101-DCN&36&$48.9$&$68.3$&$53.4$&$29.9$&$50.9$&$62.4$&CVPR 2021\\
    &Dynamic DETR \cite{DynamicDETR}&ResNeXt-101-DCN&36&$49.3$&$68.4$&$53.6$&$30.3$&$51.6$&$62.5$&ICCV 2021\\
    &Deformable DETR \cite{DDETR}&ResNeXt-101-DCN&50&$50.1$&$69.7$&$54.6$&$30.6$&$52.8$&$64.7$&ICLR 2021\\
    \hline
    \multirow{2}{*}{\scriptsize{\rotatebox{90}{\textit{Ours}}}} 
    &Corr-Sparse R-CNN&ResNet-101-DCN&36&$49.6$&$67.8$&$54.1$&$29.2$&$52.3$&$64.9$&\\  
    &Corr-Sparse R-CNN&ResNeXt-101-DCN&36&$\mathbf{51.0}$ &$69.2$ &$\mathbf{55.7}$ &$31.1$ &$\mathbf{53.7}$ &$\mathbf{66.3}$ &\\
    \hline
    \end{tabular}
            \vspace{-1ex}
    \caption{SOTA comparison on COCO \textit{test-dev}. Our Corr-Sparse R-CNN (i) performs on-par or better compared to recent  NMS-based methods, all of which also enforce correlation, and (ii) outperforms NMS-free methods by a notable margin. Results are obtained from papers.}
    \label{tab:Detectiontestdev}
            \vspace{-1ex}
\end{table*}
\subsection{Comparison with Methods Enforcing Correlation}
\label{subsec:ExpMethodswithCorr}
Table \ref{tab:enforceCorr} compares Corr. Loss. with using an aux. head \cite{ATSS}, QFL \cite{GFL} and RS Loss \cite{RSLoss} on the common ATSS baseline \cite{ATSS} wrt. detection and correlation:

\textbf{Detection Performance.} Reaching $39.8 \mathrm{AP_C}$ without an aux. head, Concordance Loss (Table \ref{tab:enforceCorr}) outperforms using an aux. head, which introduces additional learnable parameters ($39.8$ vs $39.3 \mathrm{AP_C}$), and reaches on-par performance with the recently proposed, relatively complicated loss functions, QFL \cite{GFL} and RS Loss \cite{RSLoss}. Besides, owing to its simple usage, Concordance Loss is complementary to existing methods: It yields $40.0 \mathrm{AP_C}$ with an aux. head (+0.7 $\mathrm{AP_C}$) and $40.2 \mathrm{AP_C}$ with RS Loss (+0.3 $\mathrm{AP_C}$) without introducing additional learnable parameters.

\textbf{Correlation Analysis.} To provide insight, we report $\beta_{img}$ (Eq. \ref{eq:pergt}) and $\beta_{cls}$ (Eq. \ref{eq:acrgt}) in Table \ref{tab:enforceCorr}: Our Concordance Loss (i) improves baseline correlation significantly, enhancing $\beta_{img}$ (from $27.3 \%$ to $31.6\%$) and $\beta_{cls}$ (from $40.3 \%$ to $45.2\%$) both by $\smallsim 5 \%$, and (ii) results in better correlation than all methods wrt. $\beta_{img}$ and $\beta_{cls}$ once combined with QFL and RS Loss respectively. This set of results confirms that Concordance Loss improves correlation between classification and localization tasks in both image-level and class-level.



\subsection{Comparison with SOTA}
Here, we prefer Sparse R-CNN owing to its competitive detection performance and our large gains. We train our ``Corr-Sparse R-CNN'' for 36 epochs with DCNv2 \cite{DCNv2} and multiscale training by randomly resizing the shorter side within [480, 960]  similar to common practice \cite{RSLoss,varifocalnet,sparsercnn}. Table \ref{tab:Detectiontestdev} presents the results on COCO test-dev \cite{COCO}: 

\textbf{NMS-based Methods.} On the common ResNet-101-DCN backbone and with similar  data augmentation, our Corr-Sparse R-CNN yields $49.6 \mathrm{AP_C}$ at $13.7$ fps (on a V100 GPU) outperforming recent NMS-based methods, all of which also enforce correlation, e.g., (i) RS-R-CNN \cite{RSLoss} by $1.8 \mathrm{AP_C}$, (ii) GFLv2 \cite{GFLv2} by more than $1 \mathrm{AP_C}$, and (iii) VFNet \cite{varifocalnet} in terms of not only $\mathrm{AP_C}$ but also efficiency (with $12.6$ fps on a V100 GPU).
On ResNeXt-101-DCN, our Corr-Sparse R-CNN provides $51.0 \mathrm{AP_C}$ at $6.8$ fps, surpassing all methods including RS-Mask R-CNN+ ($50.2 \mathrm{AP_C}$ at $6.4$ fps), additionally using masks and Carafe FPN \cite{carafe}.

\textbf{NMS-free Methods.} Our Corr-Sparse R-CNN outperforms (i) TSP R-CNN \cite{TSPRCNN} by more than $2 \mathrm{AP_C}$ on ResNet-101-DCN with significantly less training, (ii) Sparse R-CNN \cite{sparsercnn} by $\smallsim 2 \mathrm{AP_C}$ and Deformable DETR \cite{DDETR}, a recent strong NMS-free method, by $\smallsim 1 \mathrm{AP_C}$ on ResNeXt-101-DCN.


\subsection{Ablation \& Hyper-parameter Analyses}
\label{subsec:ablation}
\textbf{Optimizing Different Correlation Coefficients.} 
Spearman Loss yields better localization performance, i.e. the lowest localization error wrt. $\mathrm{oLRP_{Loc}}$ in all experiments while it rarely yields the best $\mathrm{oLRP_{FP}}$ or $\mathrm{oLRP_{FN}}$, implying its contribution to classification to be weaker than Concordance Loss (see Appendix for components of $\mathrm{oLRP}$). We also tried Pearson Correlation Coefficient on ATSS and Sparse R-CNN but it performed worse compared to either using Spearman or Concordance (Appendix).

\textbf{Backpropagating Through Different Heads.} On Sparse R-CNN, we observed that the performance degrades when we backpropagate either only localization head ($37.5$ AP) or both heads ($38.9$ AP). Hence, we preferred backpropagating the gradients only through the classification head ($39.3$ AP).

\textbf{Effect on Training Time.} Using Spearman or Concordance Loss to train Sparse R-CNN, computing the loss for 6 times each iteration, increases iteration time 0.50 sec to 0.51 sec on V100 GPUs, suggesting a negligible overhead.

\textbf{Sensitivity to $\lambda_{corr}$.} We found it sufficient to search over $\{0.1, 0.2, 0.3. 0.4, 0.5, 0.6\}$ to tune $\lambda_{corr}$. Appendix presents empirical results for grid search. 



\subsection{Additional Material}

This paper is accompanied by an Appendix containing (i) the effect of Corr.Loss on Sparse R-CNN using its stronger setting, (ii) components of oLRP for detectors in Table \ref{tab:nmseval}, (iii) results when Pearson Correlation Coefficient is optimized, (iv) our grid search to tune $\lambda_{corr}$.

\section{Conclusion}
\label{sec:conclusion}
In this paper, we defined measures to evaluate object detectors wrt. correlation, provided analyses on several methods and proposed Correlation Loss as an auxiliary loss function to enforce correlation for object detectors. Our extensive experiments on six detectors show that Correlation Loss. consistently improves the detection and correlation performances, and reaches SOTA results. 



\section*{Acknowledgments}
This work was supported by the Scientific and Technological Research Council of Turkey (TÜBİTAK) (under grant 120E494). We also gratefully acknowledge the computational resources kindly provided by TÜBİTAK ULAKBIM High Performance and Grid Computing Center (TRUBA) and METU Robotics and Artificial Intelligence Center (ROMER). Dr. Akbas is supported by the ``Young Scientist Awards Program (BAGEP)'' of Science Academy, Turkey.


\bibliography{aaai23}

\section*{APPENDIX}

\renewcommand{\thetable}{A.\arabic{table}}
\renewcommand{\theequation}{A.\arabic{equation}}
\renewcommand{\thesection}{A}

\begin{table}[ht!]
    \centering
    \setlength{\tabcolsep}{0.1em}
    \small
    \begin{tabular}{|c|c|c||c c c c c c|} \hline
        Method & Dataset & $0.0$&$0.1$&$0.2$&$0.3$&$0.4$&$0.5$ &$0.6$ \\ \hline\hline 
        ATSS &  COCO& $38.7$&$38.8$&$39.3$&$\mathbf{39.8}$&$39.7$&$39.7$ & $39.6$\\ 
        YOLACT & COCO&$28.3$&$28.6$&$28.8$&$28.8$&$\mathbf{29.0}$&$28.8$ &$28.6$ \\  
        Sparse R-CNN & COCO&$37.7$&$38.7$&$\mathbf{39.3}$&$39.1$&$39.0$&$38.1$ & $38.0$\\ 
        Sparse R-CNN & Cityscapes &$39.0$&$39.0$&$38.3$&$39.9$&$40.0$&$40.0$ & $\mathbf{40.8}$\\ \hline
    \end{tabular}
    \caption{Grid search to tune $\lambda_{corr}$ on different models. We present the results for concordance correlation coefficient for ATSS and YOLACT, and spearman correlation coefficient for Sparse R-CNN models. 0.0 corresponds to not including our Correlation Loss.}
    \label{tab:grid}
\end{table}

\textbf{Sensitivity to $\lambda_{corr}$.} In Table \ref{tab:grid}, we see that (i) $\lambda_{corr}=0.2$ provides the best performance overall, (ii) the performance is not very sensitive to  $\lambda_{corr}$ and (iii) a grid search over $\{0.1, 0.2, 0.3. 0.4, 0.5, 0.6\}$ is sufficient (outside of this range, performance drops).

\begin{table}
    \centering

    \begin{tabular}{|c|c|c|c|} \hline
    Method&$\mathrm{AP}$&$\mathrm{AP_{50}}$&$\mathrm{AP_{75}}$ \\ \hline\hline
    Sparse R-CNN \cite{sparsercnn} &$45.0$&$64.1$&$48.9$\\
     w. Conc.Corr (Ours)&$45.5$&$\mathbf{64.4}$&$49.7$\\
     w. Spear.Corr (Ours)&$\mathbf{46.1}$ &$64.0$&$\mathbf{50.4}$\\ \hline
    \end{tabular}
    \caption{Comparison with stronger Sparse R-CNN.}
    \label{tab:strongersetting}
\end{table}

\textbf{The effect of Corr.Loss on Sparse R-CNN using its stronger setting.} Following Sun et al. \cite{sparsercnn} (Table \ref{tab:strongersetting}), we train Sparse R-CNN with 36 epochs training, 300 proposals, multi-scale training and random cropping. Table \ref{tab:strongersetting} presents that the improvement of our Spearman Loss on this strong baseline is $\sim 1$ AP points.

\textbf{Using Pearson Correlation Coefficient.} We tried optimizing pearson correlation coefficient as well and observed that while it has similar performance with concordance correlation coefficient on ATSS and spearman correlation coefficient on Sparse R-CNN, it does not outperform the other two in both of the cases (Table \ref{tab:pearson2}). Considering the similarities of spearman and concordance correlation coefficients in terms of scoring the relation of the values, we preferred concordance correlation coefficient over spearman correlation coefficient due to the fact that concordance correlation coefficient enforces the scores to be equal to the IoUs imposing a tighter constraint than pearson correlation coefficient.

\textbf{The components of oLRP.} Table \ref{tab:nmseval_app} shows the the components of oLRP for different detectors corresponding to Table 3 in the paper. As discussed in the paper, Spearman Loss yields better localization performance, i.e. the lowest localization error wrt. $\mathrm{oLRP_{Loc}}$ in all experiments while it rarely yields the best $\mathrm{oLRP_{FP}}$ or $\mathrm{oLRP_{FN}}$, implying its contribution to classification to be weaker than Concordance Loss.

\blockcomment{
\begin{table}[h]
    \centering
    
    \begin{tabular}{|c|c|c|c|} \hline
        $\lambda_{corr}$&$\mathrm{AP_{C}}$&$\mathrm{AP_{50}}$&$\mathrm{AP_{75}}$ \\ \hline \hline
        0.0 (baseline) & $39.3$&$\mathbf{57.4}$&$42.5$ \\ \hline \hline
        0.1 &$\mathbf{39.5}$&$57.1$&$\mathbf{42.8}$ \\ \hline
        0.2 &$39.1$&$55.6$&$42.5$ \\ \hline 
        0.3 &$38.7$&$54.7$&$42.2$ \\ \hline 
        0.4 &$37.7$&$53.0$&$41.5$ \\ \hline
        0.5 &$36.7$&$51.6$&$40.2$ \\ \hline
    \end{tabular}
    \caption{ATSS with Pearson Loss}
    \label{tab:pear}
\end{table}
}

\begin{table}[h]
    \centering
    \begin{tabular}{|c|c|c|c|} \hline
        Method &$\mathrm{AP_{C}}$&$\mathrm{AP_{50}}$&$\mathrm{AP_{75}}$ \\ \hline \hline
        ATSS w/o aux head & $38.7$&$57.6$&$41.5$\\ 
        w. Pearson Corr &$39.4$&$56.6$&$42.7$ \\ 
        w. Conc.Corr &$\mathbf{39.8}$&$\mathbf{57.9}$&$\mathbf{43.2}$ \\ 
        w. Spear.Corr &$39.3$&$56.6$&$42.5$ \\ \hline 
        Sparse-RCNN & $37.7$&$55.9$ &$40.5$ \\ 
        w. Pearson Corr &$\mathbf{39.3}$&$56.6$&$42.2$ \\ 
        w. Conc.Corr &$38.9$&$\mathbf{57.2}$&$41.8$ \\ 
        w. Spear.Corr &$\mathbf{39.3}$&$56.7$&$\mathbf{42.5}$ \\ \hline 
    \end{tabular}
    \caption{Effect of using Pearson correlation coefficient.}
    \label{tab:pearson2}
\end{table}

\begin{table*}[hbt!]
    \centering
    \small
    \setlength{\tabcolsep}{0.5em}
    
    \resizebox{\textwidth}{!}{\begin{tabular}{|c||c|c|c||c|c|c|c|} \hline
         Method & $\mathrm{AP_C} \uparrow$ & $\mathrm{AP_{50}} \uparrow$ & $\mathrm{AP_{75}} \uparrow$ & $\mathrm{oLRP} \downarrow$& $\mathrm{oLRP_{Loc}} \downarrow$& $\mathrm{oLRP_{FP}} \downarrow$& $\mathrm{oLRP_{FN}} \downarrow$ \\ \hline\hline
    Retina Net \cite{FocalLoss}&$36.5$&$55.4$&$39.1$&$70.7$ &$16.8$ &$32.0$ &$\mathbf{48.1}$\\
    w. Conc.Corr (Ours) &$37.0$&$\mathbf{55.7}$&$39.7$ &$70.2$ &$16.3$ &$\mathbf{30.8}$ &$49.3$ \\ 
    w. Spear.Corr (Ours) &$\mathbf{37.5}$&$55.4$&$\mathbf{40.5}$& $\mathbf{69.7}$ &$\mathbf{16.0}$ &$31.3$ &$48.4$ \\ \hline
    Fovea Box \cite{FoveaBox}&$36.4$&$56.5$&$38.6$&$70.2$ &$17.0$ &$30.2$ &$\mathbf{47.2}$ \\
    w. Conc.Corr (Ours) &$\mathbf{37.1}$&$\mathbf{56.4}$&$\mathbf{39.6}$&$\mathbf{69.7}$ &$16.6$ &$\mathbf{28.6}$ &$48.1$\\ 
    w. Spear.Corr (Ours) &$37.0$&$55.6$&$39.3$& $70.0$ &$\mathbf{16.3}$ &$31.0$ &$47.9$\\ \hline
    ATSS \cite{ATSS} &$38.7$&$57.6$&$41.5$&$69.0$&$16.0$ &$\mathbf{29.1}$&$47.0$  \\
    w. Conc.Corr (Ours)&$\mathbf{39.8}$&$\mathbf{57.9}$&$\mathbf{43.2}$&$\mathbf{68.2}$&$15.4$ &$\mathbf{29.1}$&$46.9$ \\ 
    w. Spear.Corr (Ours) &$39.3$&$56.6$&$42.5$&$68.7$ &$\mathbf{15.2}$&$31.2$&$\mathbf{46.7}$  \\ \hline
    PAA \cite{paa} &$39.9$&$57.3$&$43.4$ &$68.6$ &$15.0$ &$30.4$  &$47.0$ \\
    w. Conc.Corr (Ours) &$\mathbf{40.7}$&$\mathbf{58.8}$&$\mathbf{44.3}$&$\mathbf{67.7}$  &$15.2$ &$\mathbf{28.5}$  &$\mathbf{46.3}$ \\ 
    w. Spear.Corr (Ours) &$40.4$&$58.0$&$43.7$&$67.8$  &$\mathbf{14.9}$  &$29.5$  &$46.6$  
\\   \hline \hline
         Sparse R-CNN \cite{sparsercnn}&$37.7$&$55.8$&$40.5$&$69.5$ &$16.0$ &$28.7$ &$48.6$\\
    w. Conc.Corr (Ours) &$38.9$&$\mathbf{57.2}$&$41.8$&$\mathbf{68.1}$ &$15.7$ &$27.7$ &$\mathbf{47.2}$ \\ 
    w. Spear.Corr (Ours) &$\mathbf{39.3}$ &$56.7$ &$\mathbf{42.5}$ &$68.3$ &$\mathbf{15.3}$ &$\mathbf{27.1}$ &$48.4$ \\ \hline
    
    \end{tabular}}
    \caption{Components of oLRP for Table 3 in the paper.}
    \label{tab:nmseval_app}
\end{table*}

\end{document}